%% file: Teleoperation.tex
\documentclass[lettersize,journal]{IEEEtran}
\usepackage{amsmath,amsfonts}
\usepackage{algorithmic}
\usepackage{algorithm}
\usepackage{array}
\usepackage[caption=false,font=normalsize,labelfont=sf,textfont=sf]{subfig}
\usepackage{textcomp}
\usepackage{stfloats}
\usepackage{url}
\usepackage{verbatim}
\usepackage{amssymb}
\usepackage{graphicx}
\usepackage[utf8]{inputenc}
\usepackage[T1]{fontenc}
\usepackage{wrapfig}

\usepackage{amsmath}
\usepackage{multirow}
\usepackage[nottoc]{tocbibind}
\usepackage{float}
\usepackage{soul}
\usepackage[justification=centering]{caption}
\usepackage[english]{babel}
\usepackage{colortbl}
\usepackage{amsfonts}
\usepackage{graphicx}
\usepackage{lipsum}
\usepackage{multicol}

\usepackage{booktabs}
\usepackage[referable]{threeparttablex}
\usepackage{algorithm}
\usepackage{algorithmic}
\usepackage[dvipsnames]{xcolor}
\definecolor{objcolor}{rgb}{204,204,255}
\usepackage{subfiles}
\usepackage{nomencl}
\sethlcolor{yellow}
\usepackage{xcolor}

\usepackage{url}
\makenomenclature

\newcolumntype{L}{>{\centering\arraybackslash}m{3cm}}
\newcolumntype{S}{>{\centering\arraybackslash}m{1.5cm}}

\begin{document}

\title{A Neural Network Based Teleoperation for Remote Controlled Vehicles}


\author{Ning Ding, Azim Eskandarian \textit{IEEE Fellow}

\thanks{This work has been accepted by IEEE Transactions on Control Systems Technology.}
\thanks{Ning Ding was with Virginia Polytechnic Institute and State University, Blacksburg, VA 24060 USA (e-mail: ningding@vt.edu)}
\thanks{Azim Eskandarian is with College of Engineering, Virginia Commonwealth University, Richmond, VA 23284 USA (e-mail: eskandariana@vcu.edu)}
}



\maketitle

\subfile{Abstract}  

\subfile{Introduction}
\subfile{Method}

\subfile{Simulation}
\subfile{Experiment}
\subfile{CONCLUSIONS_and_FUTURE_WORK}

\bibliographystyle{unsrt}
\bibliography{ref}

\addtolength{\textheight}{-12cm}

\subfile{Biography}
\end{document}

%% file: Abstract.tex
\begin{abstract}

Direct teleoperation of vehicles faces critical technical bottlenecks: communication latency and the operator's inability to physically perceive unmodeled environmental disturbances (e.g., aerodynamic drag, bank angles) coupled with highly nonlinear tire-road dynamics. 
To address these challenges, we propose a tailored unilateral teleoperation framework. 
The system integrates the Wave Variable (WV) approach to passively guarantee stability under stochastic delays, and an adaptive Radial Basis Function Network (RBFN) to actively compensate for vehicle-specific uncertainties.
Unlike existing WV-neural network architectures designed for bilateral robotic arms, our framework features decoupled adaptive laws specifically designed for vehicle longitudinal and lateral dynamics. 
Furthermore, compared to model-heavy predictive controllers, the model-free RBFN offers rapid online adaptation without heavy computational overhead. 
Building upon our preliminary theoretical formulation, this brief paper presents comprehensive comparative analyses and real-world hardware validations. 
Simulation benchmarks against PID, LQR, MPC, and NMPC demonstrate that the RBFN achieves superior robustness against unmodeled disturbances while requiring orders of magnitude less execution time than MPC and NMPC, making it ideal for resource-constrained vehicle edge computing. 
Finally, hardware-in-the-loop experiments using a 1/10th scale vehicle over a 4G network validate the system's practical feasibility, safety, and robust trajectory tracking under physical road uncertainties.

\end{abstract}

\begin{IEEEkeywords}
Teleoperation, Remote Controlled Vehicle, Wave Variable Approach, Radial Basis Function Network, Longitudinal Control, Lateral Control
\end{IEEEkeywords}

%% file: Introduction.tex
\section{Introduction}

Several U.S. states have legalized autonomous-vehicle teleoperation \cite{legaltele2021}, underscoring its role as a human-in-the-loop safety fallback. Teleoperation enables remote intervention when onboard autonomy encounters situations outside its operational design domain, supporting practical deployment.

Most prior work emphasizes monitoring or indirect control, while direct vehicle teleoperation (continuous remote steering/throttle authority) remains less explored. Surveys such as Majstorović et al. \cite{9945267} summarize direct-control concepts, but two core issues remain insufficiently addressed: communication latency and unmodeled vehicle dynamics/disturbances \cite{prakash2022vehicle}. These gaps motivate this work.

To mitigate latency, predictive approaches such as MPC \cite{hatori2021teleoperation} and NMPC \cite{surjana2024application} are commonly used. In teleoperation, however, their effectiveness depends on accurate plant models; sudden, unmodeled disturbances during delay intervals can severely degrade prediction quality. NMPC also incurs substantial computational cost, which is unfavorable for real-time execution on vehicle edge hardware.

An alternative is the Wave Variable (WV) approach \cite{niemeyer1991stable, li2013achieving, sun2014application, alise2009extending}, which provides a passive, model-free stability guarantee under arbitrary delays. While WV is mature in teleoperated robotic manipulators \cite{aziminejad2008transparent,lee2006passive}, applying it to vehicles is not straightforward. Manipulator teleoperation is typically bilateral (force feedback), whereas vehicle teleoperation is generally unilateral; the operator does not require disruptive steering-torque reflection. This calls for a vehicle-specific WV signal architecture that encodes relevant motion states (e.g., velocity, steering angle, yaw rate) without reflecting environmental resistance to the operator.

WV alone does not actively compensate for unmodeled dynamics and disturbances, for which neural networks are effective. 
Prior WV--NN designs (e.g., WV with Radial Basis Function Networks (RBFNs) \cite{7593387}) target bilateral robotic arms and do not directly transfer to unilateral vehicle teleoperation. 
Here, we propose a unilateral, vehicle-oriented WV framework combined with Lyapunov-based, decoupled longitudinal/lateral RBFN adaptation for direct road-vehicle teleoperation under stochastic delays and environmental disturbances. 
A single-layer RBFN is adopted instead of multi-layer neural networks \cite{jagannathan2001control,dixon1992neural} to approximate low-dimensional lumped uncertainties driven by the tracking-error vectors $E_1$ and $E_2$. 
This choice enables efficient real-time adaptation and facilitates a Lyapunov-based adaptive law design with guaranteed closed-loop stability.

A preliminary theoretical formulation, including basic Lyapunov analysis, appeared in \cite{DING2024168}. This paper extends that work with experimental validation and comparative benchmarking. Our main contributions are:
\begin{itemize}
\item We benchmark the proposed controller against PID, LQR, MPC, and NMPC, showing improved robustness to unmodeled disturbances while requiring orders-of-magnitude less execution time than MPC and NMPC.
\item We validate the framework on a 1/10th-scale remote vehicle (RV) operated via a gaming steering wheel over a real 4G network with stochastic latency, demonstrating practical robustness under physical uncertainties and operator variability.
\end{itemize}

%% file: Method.tex
\section{METHOD}
\label{sec:method}
\subfile{pic/big_pic}

Figure~\ref{fig:big_pic} shows the proposed teleoperation framework, which comprises (i) a communication module, (ii) a trajectory design module, and (iii) an adaptive RBFN-based controller. Communication robustness to stochastic delays is achieved using WV, while closed-loop stability and online uncertainty compensation are ensured via Lyapunov-based adaptive laws.

\subsection{Communication Module (Wave Variable Signal Transmission)}
\label{sec:wv}

Vehicle teleoperation is unilateral: the operator relies on visual
  feedback and does not require force/torque reflection, which could
  inject road disturbances into the steering interface. We therefore
  employ the WV framework~\cite{niemeyer1991stable} to passively transmit
  command/state signals, ensuring stability under time-varying delays
  without reflecting environmental forces.

  Two WV channels carry the power-conjugate pairs
  $(x_{*1},y_{*1})\!=\!(F_*,v_*)$ and
  $(x_{*2},y_{*2})\!=\!(\dot\delta_*,\varphi_*)$, where $*\!\in\!\{l,f\}$
  denotes the operator (leader) and the RV (follower). For each channel
  $i\!\in\!\{1,2\}$, the wave variables are
  \begin{equation}\label{equ:wv}
  u_{*i}=\frac{k_ix_{*i}+y_{*i}}{\sqrt{2k_i}},\qquad
  w_{*i}=\frac{k_ix_{*i}-y_{*i}}{\sqrt{2k_i}},
  \end{equation}
  with impedances $k_i\!>\!0$. Forward/backward transmission with delays
  $T_1,T_2$ gives
  \begin{equation}\label{equ:wvdelay}
  u_{fi}(t)=u_{li}(t-T_1),\qquad w_{li}(t)=w_{fi}(t-T_2).
  \end{equation}
  Decoding recovers the commands and states as
  \begin{equation}\label{equ:wvdecode}
  x_{fi}=\frac{u_{fi}+w_{fi}}{\sqrt{2k_i}},\qquad
  y_{li}=\sqrt{\tfrac{k_i}{2}}\,(u_{li}-w_{li}),
  \end{equation}
  yielding $(F_f,\dot\delta_f)$ at the RV and $(v_l,\varphi_l)$ at the
  operator.

Packet loss can be modeled in the WV framework as intermittent signal interruptions. Although such losses may degrade tracking performance, passivity can still be preserved with an appropriate loss-handling strategy. Because latency is the dominant network impairment considered here, we focus on delay effects; a quantitative analysis of packet-loss impacts is left for future work.

\subsection{Trajectory Generation and Vehicle Model}
The decoded commands are applied to a standard bicycle kinematic model~\cite{DING2024168} to generate the reference trajectory for the RV, including desired position $(X_d,Y_d)$, velocity components $(v_{xd},v_{yd})$, yaw angle $\varphi_d$, and yaw rate $r_d$.

For tracking, we consider the RV longitudinal and lateral dynamics. The longitudinal dynamics are:
\begin{equation}
   \dot{v}_{xf}=\frac{1}{M}\Big(\frac{1+\cos\delta_f}{2}F_c-F_{yf}\sin\delta_f-F_{res}+Mv_{yf}r_f\Big)
    \label{equ:longdyna}
\end{equation}
where $M$ represents RV mass; $F_{res}=F_r+F_a+F_g$; $F_c$ denotes the control (command) force; $v_{xf}$ and $v_{yf}$ are the longitudinal and lateral velocities, respectively; $r_f$ is the yaw rate; $\delta_f$ is the steering angle; $F_{yf}$ is the front lateral tire force; $F_r$, $F_a$, and $F_g$ represent the rolling resistance, aerodynamic drag, and grade resistance due to road slope, respectively.

The lateral dynamics are modeled as:
\begin{equation}
   \begin{bmatrix}\ddot{Y}_f\\ \ddot{\varphi}_f\end{bmatrix}
=H\begin{bmatrix}\dot{Y}_f\\ \dot{\varphi}_f\end{bmatrix}+D\delta_c +\begin{bmatrix}
F_\beta\\ 0\end{bmatrix} 
    \label{equ:latdyn}
\end{equation}
\begin{equation}
   H = \begin{bmatrix}
-\frac{2C_f+2C_r}{Mv_{xf}} &-Mv_{xf}-\frac{2C_fl_f-2C_rl_r}{Mv_{xf}} \\ 
-\frac{2C_fl_f-2C_rl_r}{I_zv_{xf}} & -\frac{2C_fl_f^2+2C_rl_r^2}{I_zv_{xf}} 
\end{bmatrix}
    \label{equ:aaaaaa1}
\end{equation}
\begin{equation}
   D=\begin{bmatrix}\frac{2C_f}{M} & \frac{2C_fl_f}{I_z}\end{bmatrix}^T
    \label{equ:ddddd}
\end{equation}
where $I_z$ is the RV yaw moment of inertia; $C_f$ ($C_r$) denotes the front (rear) tire cornering stiffness; $l_f$ ($l_r$) is the distance from the RV's center of mass to the front (rear) axle; $F_\beta$ is the lateral force induced by the road bank angle; $Y_f$ is RV's lateral position; $\varphi_f$ is RV's yaw angle; and $\delta_c$ is the controlled steering input.

\subsection{Adaptive RBFN Controller}
\label{sec:rbfn_controller}
The WV guarantees passivity under delays, but does not address modeling errors, environmental resistance, and lateral nonlinearities. We therefore add an RBFN as an online compensation term.

A single-layer RBFN is used:
\begin{equation}
    h_j(z)=\exp\!\Big(-\frac{\lVert z-c_j\rVert^2}{b_j^2}\Big),\qquad f_i=W_i^T h(z)
\label{equ:rbfn1}
\end{equation}
where $z$ is the input vector, $c_j$ and $b_j$ are the center and width of node $j$, and $W_i$ denotes the weights for output $i$. 

The centers are placed on a uniform grid over the input operating
    box for $z$, which is estimated from the admissible tracking-error
    envelope and the desired-trajectory range. With the Gaussian form
    used in {\eqref{equ:rbfn1}}, the widths are set by the neighbor-overlap rule $b_j = d_{\mathrm{eff}}/\sqrt{\ln(1/\rho)}$, where
    $d_{\mathrm{eff}}$ is the grid spacing and $\rho \in (0,1)$ is the
    value of $h_j$ evaluated at an adjacent center; we use $\rho = 0.5$
    so that each basis function decays to half of its peak at the
    location of its nearest neighbors. The resulting grid size, grid
    spacing and $\rho$ are chosen once from the operating range and kept
    fixed across all experiments reported below, and the Lyapunov-based adaptive law ensures bounded tracking behavior under the assumed RBFN approximation conditions.

\subsubsection{Longitudinal Control}
Define the longitudinal tracking error vector $E_1=\begin{bmatrix}e_1 & \dot{e}_1\end{bmatrix}^T$. The RBFN estimates lumped longitudinal uncertainty:
\begin{equation}
    \hat{f}=W_{1}^{T}h(E_1)
    \label{equ:hatf}
\end{equation}
The control law is:
\begin{equation}
    F_c=\frac{2}{1+\cos\delta_f}\Big(M\dot{v}_{xd}+\hat{f}-k_{p1}e_1-k_{d1}\dot{e}_1\Big)
    \label{equ:longctllaw}
\end{equation}
with gains $k_{p1},k_{d1}>0$. Using the Lyapunov framework in~\cite{DING2024168}, the adaptive weight update is:
\begin{equation}
      \dot{W}_1=-\gamma E_1^TP_1Bh(E_1)
      \label{equ:laonglaw}
\end{equation}
where $\gamma>0$, $B=\begin{bmatrix}0 & \frac{1}{M}\end{bmatrix}^T$, and $P_1$ is symmetric positive definite.

\subsubsection{Lateral Control}
Define the lateral error vector $E_2=\begin{bmatrix}e_2 & \dot{e}_2 & e_3 & \dot{e}_3\end{bmatrix}^T$. The RBFN estimates lumped lateral nonlinearities:
\begin{equation}
    \hat{g}=W_{2}^{T}h(E_2)
    \label{equ:hatg}
\end{equation}
The steering control is:
\begin{equation}
\delta_c =D^{\dagger }\Big(\begin{bmatrix}\ddot{Y}_{d}\\ \ddot{\varphi}_{d}\end{bmatrix}-\hat{g}\Big)
-k_{p2}e_2-k_{d2}\dot{e}_2-k_{p3}e_{3}-k_{d3}\dot{e}_3
\label{equ:latctllaw}
\end{equation}
where $D^{\dagger}$ is the pseudo-inverse of $D$ and $k_{p2},k_{d2},k_{p3},k_{d3}>0$. The Lyapunov-based adaptive update~\cite{DING2024168} is:
\begin{equation}
    \dot{W}_2=\eta h(E_2)E_2^TP_2L 
    \label{equ:laterallawa}
\end{equation}
where $\eta>0$, $P_2$ is symmetric positive definite, and
$L=\begin{bmatrix}0 &0&0&1 \\ 0&1&0&0\end{bmatrix}^T$.

 \subsection{Stability Analysis of the Delayed WV--RBFN Closed Loop}
\label{sec:stable}

We show that the composite tracking error
  $E=[E_1^{\mathsf T},E_2^{\mathsf T}]^{\mathsf T}\!\in\!\mathbb R^{6}$
  is uniformly ultimately bounded (UUB) under arbitrary bounded delays
  $T_1,T_2\!\ge\!0$, given (i) bounded desired trajectories and decoded
  port signal $x_f$, (ii) bounded Gaussian basis $\|h(\cdot)\|\!\le\!h_M$,
  and (iii) a passive operator
  $\dot S_{\mathrm{op}}\!\le\!-x_l^{\mathsf T}y_l$.

  \paragraph{Step 1: Wave-variable channel passivity.}
  With port signals
  $x_l\!=\![F_l,\dot\delta_l]^{\mathsf T}$,
  $y_l\!=\![v_l,\varphi_l]^{\mathsf T}$,
  $x_f\!=\![F_f,\dot\delta_f]^{\mathsf T}$,
  $y_f\!=\![v_f,\varphi_f]^{\mathsf T}$,
  and storage
  $S_{WV}(t)\!=\!\tfrac12\!\sum_{i=1}^{2}\!\bigl[\!\int_{t-T_1}^{t}\!u_{li}^{2}d\tau
  +\!\int_{t-T_2}^{t}\!w_{fi}^{2}d\tau\bigr]$,
  the wave transformation \cite{niemeyer1991stable} yields
  \begin{equation}\label{equ:wvpassivity}
  \dot S_{WV}\le x_l^{\mathsf T}y_l-x_f^{\mathsf T}y_f,
  \end{equation}
  independent of $T_1,T_2$.

  \paragraph{Step 2: Vehicle Lyapunov bound}
  Under the ideal RBFN identities adopted in \cite{DING2024168}, the
  closed loops reduce to
  \begin{equation}\label{eq:cl}
  \dot E_1=A_1E_1+B\widetilde W_1^{\mathsf T}h(E_1),\quad
  \dot E_2=A_2E_2+L\widetilde W_2^{\mathsf T}h(E_2),
  \end{equation}
  with $\widetilde W_j=W_{oj}-W_j$ and matrices $A_1,A_2$
  determined by the design gains (explicit forms in \cite{DING2024168}).
  With the composite Lyapunov function
  \begin{equation}\label{eq:VRV}
  V_{\mathrm{RV}}=\tfrac12 E_1^{\mathsf T}P_1E_1+\tfrac{1}{2\gamma}\widetilde W_1^{\mathsf T}\widetilde W_1
  +\tfrac12 E_2^{\mathsf T}P_2E_2+\tfrac{1}{2\eta}\mathrm{tr}(\widetilde W_2^{\mathsf T}\widetilde W_2),
  \end{equation}
  $P_j\!\succ\!0$ satisfying $A_1^{\mathsf T}P_1+P_1A_1=-Q_1$,
  $A_2^{\mathsf T}P_2+P_2A_2=-Q_2$, and the adaptive laws
  $\dot W_1=-\gamma E_1^{\mathsf T}P_1Bh(E_1)$,
  $\dot W_2=\eta h(E_2)E_2^{\mathsf T}P_2L$,
  \cite{DING2024168} establishes
  $\dot V_j=-\tfrac12 E_j^{\mathsf T}Q_jE_j$ for $j=1,2$. Summing and
  applying the Rayleigh inequality yields
  \begin{equation}\label{eq:vehbound}
  \dot V_{\mathrm{RV}}\le -\alpha_0\|E\|^{2},\qquad
  \alpha_0=\tfrac12\min\!\bigl(\lambda_{\min}(Q_1),\lambda_{\min}(Q_2)\bigr),
  \end{equation}
  where $\lambda_{\min}(\cdot)$ denotes the smallest eigenvalue of a
  symmetric positive-definite matrix.

  \paragraph{Step 3: Composite storage and UUB.}
  Because the trajectory module integrates $x_f$ to $(v_{xd},\varphi_d)$,
  the follower port output decomposes as
  \begin{equation}\label{eq:portdecomp}
  y_f=y_f^{\star}+CE,\qquad y_f^{\star}=[v_{xd},\varphi_d]^{\mathsf T},
  \end{equation}
  where $C\!\in\!\mathbb R^{2\times 6}$ extracts $(\dot e_1,e_3)$ from
  $E=[E_1^{\mathsf T},E_2^{\mathsf T}]^{\mathsf T}$. By assumption~(i),
  $|x_f^{\mathsf T}y_f^{\star}|\!\le\!\mu_0$ and
  $\|C^{\mathsf T}x_f\|\!\le\!\mu_1$ for some $\mu_0,\mu_1>0$.
  Define $S=S_{\mathrm{op}}+S_{WV}+V_{\mathrm{RV}}\!\ge\!0$. Summing the
  operator passivity, \eqref{equ:wvpassivity}, and \eqref{eq:vehbound}
  cancels $\pm x_l^{\mathsf T}y_l$; substituting \eqref{eq:portdecomp}
  and applying Young's inequality,
  \begin{equation}\label{eq:Sdotfinal}
  \dot S \le -\tfrac{\alpha_0}{2}\|E\|^{2}+\tfrac{\mu_1^{2}}{2\alpha_0}+\mu_0.
  \end{equation}
  Hence $\dot S<0$ whenever
  \begin{equation}\label{eq:UUBradius}
  \|E\|>R:=\sqrt{\frac{\mu_1^{2}}{\alpha_0^{2}}+\frac{2\mu_0}{\alpha_0}}
  \end{equation}
  so $E$ is UUB. The delays enter only through the integration limits of $S_{WV}$ 
  ; $R$ depends on operator
  amplitudes ($\mu_0,\mu_1$) and design gains ($\alpha_0$) but not on
  $T_1,T_2$.

%% file: pic/big_pic.tex
\begin{figure*}[ht]
    \centering
    \captionsetup{justification=centering}
    \includegraphics[height=4.5cm]{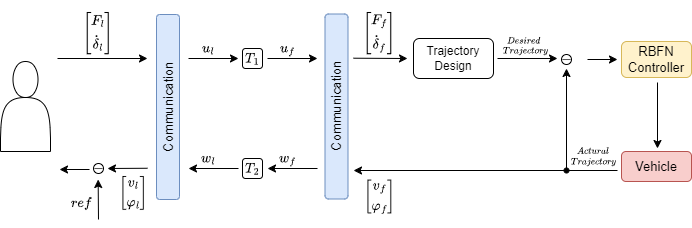}
    \caption{Architecture of the proposed neural-network--based
  teleoperation control system. The operator side (subscript $l$)
  generates command signals $(F_l, \delta_l)$ from the pedals and steering
  wheel; the remote vehicle side (subscript $f$) returns
  state signals $(v_f, \varphi_f)$. The two sides are connected
  through a wave-variable channel with forward delay $T_1$ and
  backward delay $T_2$. The adaptive RBFN controller on the vehicle side compensates online for unmodeled longitudinal and lateral
  disturbances.}
    \label{fig:big_pic}
\end{figure*}

%% file: Simulation.tex
\section{Simulation and Evaluation}
\label{sec:simulation}

We evaluate the proposed framework using a complete teleoperation system built in MATLAB/Simulink. 
The human operator is modeled as a PI controller to provide a consistent baseline. 
The RV is simulated with a 3-DOF nonlinear dynamics model \cite{mitschke1972dynamik,rajamani2011vehicle,pacejka2005tire} covering longitudinal, lateral, and yaw motions, including aerodynamic drag, rolling resistance, and road bank effects.
All controllers are sampled at $20$\,ms. When a controller
  fails to complete within one sample interval, the actuator holds the
  most recent command (zero-order hold on overrun); this gives a
  realistic real-time emulation rather than an idealized comparison.

\subsection{Compared Controllers and Implementation Details}
We benchmark the proposed RBFN controller against PID, LQR, MPC, and NMPC.
To ensure a fair comparison, each baseline controller is tuned to achieve its best performance under the same simulation/vehicle settings. Specifically, the PID gains are first initialized by the Ziegler--Nichols ultimate-sensitivity method and then refined by bounded grid search to minimize the integral absolute error on the nominal trajectory, following common practice for tracking control of autonomous vehicles \cite{farag2020pid,albhaisi2026prisma}; the LQR weighting matrices are chosen to balance tracking accuracy and control effort while maximizing closed-loop performance; and MPC/NMPC use the same vehicle kinematic/dynamic models, with a prediction horizon of 10 and a control horizon of 2, selected to provide strong tracking performance subject to computational constraints.

Unlike MPC and NMPC, the RBFN controller does not explicitly enforce input/state constraints; actuator limits are handled by low-level saturation.

\subsection{Test Conditions: Delay and Road Uncertainties}
Communication latency is modeled as a stochastic delay sampled from a Generalized Extreme Value (GEV) distribution based on 4G measurements. The nominal forward delay is about 60\,ms \cite{prakash2022vehicle}, and we additionally test forward delays up to 500\,ms to evaluate the robustness of the proposed RBFN-based controller under large delays.

Each experiment follows the same procedure: the RV tracks a predefined reference and reaches a steady target; then road disturbances are applied to evaluate disturbance rejection.

\textbf{Longitudinal tests:} target speeds are $15$, $60$, and $120\,\mathrm{km/h}$. Performance metrics are the maximum speed deviation before and after disturbances ($h_1$, $h_2$) and the corresponding settling times ($t_1$, $t_2$). Disturbances are introduced by increasing road slope to $8^\circ$ and wind speed to $10\,\mathrm{m/s}$.

\textbf{Lateral tests:} a lane-change maneuver is executed within 3\,s at $15$, $60$, and $120\,\mathrm{km/h}$. Metrics are the maximum lateral deviation from the target lane centerline before/after disturbances ($l_1$, $l_2$) and settling times ($t_1$, $t_2$). Disturbances are introduced by increasing road bank angle to $8^\circ$.

We record the per-step controller execution time, denoted $t_E$.

In addition, to assess robustness under aggressive maneuvers, we follow the ISO 3888-1 severe double lane-change procedure at high speeds of $100$ and $130\,\mathrm{km/h}$. We report the maximum lateral deviation after each lane change ($l_1$, $l_2$).

\subsection{Results and Analysis}
\subfile{tables/ad_v15}

Table \ref{Tab:ad_vel_all} shows that all controllers achieve small deviations under nominal conditions. Under road uncertainties, PID and MPC exhibit longer settling times. In particular, MPC degrades when disturbances are not captured by its internal model.
This also reveals the limitations of each baseline algorithm. PID fails to reshape system responses under unmodeled disturbances; MPC exhibits performance degradation when disturbances exceed the coverage of its internal model; while NMPC incurs excessive per-step computational overhead, making it impractical for edge deployment on vehicles with limited computing resources. Among all compared controllers, RBFN simultaneously delivers accuracy comparable to MPC, latency on par with PID, and robust online disturbance adaptation capability.

As discussed in Section \ref{sec:method}, the RBFN acts as a lumped uncertainty estimator. Using the tracking-error vector as input, it compensates unmodeled dynamics and varying resistances. 
Consequently, under wind and slope disturbances, RBFN maintains small deviations and exhibits faster recovery than MPC and NMPC, although MPC/NMPC can explicitly handle constraints that RBFN does not natively enforce.

\subfile{tables/ad_vdelay}
\subfile{tables/ad_l15}
\subfile{tables/ad_ldelay}
\subfile{tables/ad_lat_server}

Table \ref{Tab:ad_lateral_all} summarizes lateral performance at $15$, $60$, and $120\,\mathrm{km/h}$. All controllers track the lane-change trajectory satisfactorily, and RBFN performs comparably to LQR, MPC, and NMPC.

In terms of computation, RBFN achieves an execution time close to PID, while MPC and NMPC require substantially more time per step. This makes RBFN more suitable for real-time edge computing with limited resources.

Tables \ref{Tab:ad_vdelay} and \ref{Tab:ad_ldelay} further indicate that larger delays slightly increase both deviation and settling time for RBFN, but performance remains within an acceptable range.

Table {\ref{Tab:ad_serve}} shows that, for the evaluated ISO 3888-1 double lane-change scenarios at $100\,\mathrm{km/h}$ and $130\,\mathrm{km/h}$, the proposed controller achieves lateral deviations below $0.24\,\mathrm{m}$. These results indicate satisfactory lateral tracking performance under the specific test conditions considered.

%% file: tables/ad_v15.tex
\begin{table}[ht]
\centering
\scriptsize
\caption{Longitudinal Control Results for \\Different Vehicle Speeds}
\begin{tabular}{m{0.7cm}<{\centering}|m{1.0cm}<{\centering}|m{0.7cm}<{\centering}m{0.7cm}<{\centering}m{0.7cm}<{\centering}m{0.7cm}<{\centering}m{0.7cm}<{\centering}}
\toprule[1.3pt]
\midrule[0.3pt]
Speed & Controller & $t_1$ & $h_1$  & $t_2$  & $h_2$  & $t_E$ \\
(km/h) &  & (s) & (km/h) & (s) & (km/h) & (ms) \\
\midrule
\multirow{4}{*}{15} 
& RBFN & 1.3 & 0.3 & 6.5 & 1.3 & 0.03 \\
\cmidrule{2-7}
& PID & 3.7 & 0.3 & 27.5 & 2.7 & 0.02 \\
\cmidrule{2-7}
& MPC & 1.3 & 0.3 & 11.3 & 0.8 & 1.81 \\
\cmidrule{2-7}
& NMPC & 1.2 & 0.3 & 8.3 & 0.7 & 41.21 \\
\midrule
\multirow{4}{*}{60}
& RBFN & 1.5 & 0.8 & 7.6 & 1.6 & 0.03 \\
\cmidrule{2-7}
& PID & 4.6 & 1.0 & 32.2 & 3.1 & 0.01 \\
\cmidrule{2-7}
& MPC & 3.1 & 0.9 & 10.1 & 1.3 & 1.79 \\
\cmidrule{2-7}
& NMPC & 2.6 & 0.8 & 9.3 & 0.9 & 39.56 \\
\midrule
\multirow{4}{*}{120}
& RBFN & 2.1 & 1.1 & 8.8 & 1.9 & 0.03 \\
\cmidrule{2-7}
& PID & 5.8 & 1.2 & 48.9 & 3.2 & 0.02 \\
\cmidrule{2-7}
& MPC & 4.1 & 1.2 & 10.8 & 1.5 & 2.01 \\
\cmidrule{2-7}
& NMPC & 3.1 & 1.0 & 9.2 & 1.5 & 44.38 \\
\midrule
\end{tabular}
\label{Tab:ad_vel_all}
\end{table}

%% file: tables/ad_vdelay.tex
\begin{table}[!h]
\centering
\scriptsize
\caption{Results for Longitudinal Control at 60 $km/h$\\ under varying communication delays}
\begin{tabular}{m{1.2cm}<{\centering}|m{1.2cm}<{\centering}m{1.2cm}<{\centering}m{1.2cm}<{\centering}m{1.2cm}<{\centering}}

\toprule[1.3pt]
\midrule[0.3pt]
delays $(ms)$      & $t_{1}\, (s)$  & $h_{1}\, (km/h)$    & $t_{2}\, (s)$ & $h_{2}\, (km/h)$    \\ 
\midrule
300    & 2.3     & 1.0          & 8.6          & 2.1           \\ 
                         
\midrule
400     & 4.4     & 1.2          & 8.8          & 2.3           \\ 
\midrule
500     & 5.5     & 1.3         & 10.4          & 2.6           \\ 
\midrule
\end{tabular}
\label{Tab:ad_vdelay}
\end{table}

%% file: tables/ad_l15.tex
\begin{table}[!h]
\centering
\scriptsize 
\caption{Lateral Control Results for \\ Different Vehicle Speeds}
\begin{tabular}{m{0.7cm}<{\centering}|m{1.0cm}<{\centering}|m{0.7cm}<{\centering}m{0.7cm}<{\centering}m{0.7cm}<{\centering}m{0.7cm}<{\centering}m{0.7cm}<{\centering}}
\toprule[1.3pt]
\midrule[0.3pt]
Speed & Controller & $t_1$ & $l_1$  & $t_2$  & $l_2$  & $t_E$ \\
(km/h) &  & (s) & (m) & (s) & (m) & (ms) \\
\midrule
\multirow{4}{*}{15} 
& RBFN  & 15.6 & 0.15 & 22.6 & 0.08 & 0.03 \\
\cmidrule{2-7}
& LQR   & 14.3 & 0.19 & 21.5 & 0.07 & 0.12 \\
\cmidrule{2-7}
& MPC   & 13.1 & 0.18 & 19.8 & 0.07 & 2.40 \\
\cmidrule{2-7}
& NMPC  & 10.6 & 0.15 & 15.8 & 0.07 & 72.40 \\
\midrule
\multirow{4}{*}{60}
& RBFN  & 17.3 & 0.19 & 26.6 & 0.08 & 0.03 \\
\cmidrule{2-7}
& LQR   & 14.8 & 0.21 & 23.7 & 0.07 & 0.11 \\
\cmidrule{2-7}
& MPC   & 14.1 & 0.20 & 22.1 & 0.07 & 2.49 \\
\cmidrule{2-7}
& NMPC  & 11.1 & 0.15 & 16.3 & 0.06 & 75.20 \\
\midrule
\multirow{4}{*}{120}
& RBFN  & 18.4 & 0.23 & 27.3 & 0.08 & 0.03 \\
\cmidrule{2-7}
& LQR   & 16.8 & 0.26 & 25.9 & 0.08 & 0.12 \\
\cmidrule{2-7}
& MPC   & 15.9 & 0.26 & 24.2 & 0.08 & 2.58 \\
\cmidrule{2-7}
& NMPC  & 14.2 & 0.26 & 22.2 & 0.06 & 72.58 \\
\midrule
\end{tabular}
\label{Tab:ad_lateral_all}
\end{table}

%% file: tables/ad_ldelay.tex
\begin{table}[!h]
\centering
\scriptsize
\caption{Results for Lateral Control at 60 $km/h$ \\under varying communication delays}
\begin{tabular}{m{1.2cm}<{\centering}|m{1.2cm}<{\centering}m{1.2cm}<{\centering}m{1.2cm}<{\centering}m{1.2cm}<{\centering}}

\toprule[1.3pt]
\midrule[0.3pt]
delays $(ms)$     & $t_{1}\, (s)$  & $l_{1}\, (m)$    & $t_{2}\, (s)$ & $l_{2}\, (m)$    \\ 
\midrule
300   & 19.1    & 0.36          & 27.1       & 0.08       \\ 
                         
\midrule
400     & 22.1     & 0.38          & 33.6          & 0.08           \\ 
\midrule
500    & 24.5     & 0.39         & 34.2         & 0.08           \\ 
\midrule
\end{tabular}
\label{Tab:ad_ldelay}
\end{table}

%% file: tables/ad_lat_server.tex
\begin{table}[!h]
\centering
\scriptsize
\caption{Results for Lateral Deviation in ISO~3888-1 Double Lane-Change}
\begin{tabular}{m{1.2cm}<{\centering}|m{1.2cm}<{\centering}|m{1.2cm}<{\centering}m{1.2cm}<{\centering}}
\toprule[1.3pt]
\midrule[0.3pt]
Speed & Controller  & $l_1$  & $l_2$   \\
(km/h) &   & (m)  & (m)  \\
\midrule
\multirow{4}{*}{100} 
& RBFN  & 0.21 & 0.22  \\
\cmidrule{2-4}
& LQR   & 0.24 & 0.28 \\
\cmidrule{2-4}
& MPC   & 0.24 & 0.27  \\
\cmidrule{2-4}
& NMPC  & 0.23 & 0.26  \\
\midrule
\multirow{4}{*}{130}
& RBFN  & 0.23 & 0.24  \\
\cmidrule{2-4}
& LQR   & 0.27 & 0.30  \\
\cmidrule{2-4}
& MPC   & 0.26 & 0.30  \\
\cmidrule{2-4}
& NMPC  & 0.26 & 0.29  \\
\midrule

\end{tabular}
\label{Tab:ad_serve}
\end{table}

%% file: Experiment.tex
\section{Experimental Evaluation}
Section~\ref{sec:simulation} evaluates the teleoperation system in simulation using a PI controller as a simplified operator model, which provides a consistent baseline but cannot capture human prediction, adaptation, reaction delay, and nonlinear behavior. In addition, real networks introduce effects beyond pure latency (e.g., packet loss and bandwidth variation). 
Therefore, we complement the simulations with human-in-the-loop experiments conducted over an actual wireless network to validate feasibility under realistic operator variability and communication delay.

\subsection{Experimental Setup}
\subsubsection{Operator Station and Network}
The operator station uses a Logitech G29 steering wheel and pedals. 
Steering and throttle inputs are linearly mapped to the vehicle steering angle and traction-force (torque) commands. 
Visual feedback is provided by an onboard camera stream. 
The system runs on ROS2 (Foxy) at a period of $20$ ms.

Experiments are conducted over a 4G wireless network. 
The measured one-way latency is approximately 100~ms from the operator to the RV and 300~ms in the reverse direction. 
The experimental latency differs from simulation assumptions due to real network conditions.
To test more challenging conditions, we additionally emulate extended latency levels from the operator to the RV (300~ms, 400~ms, and 500~ms). 
These tests evaluate sensitivity to delay; isolating packet-loss effects would require controlled injection or precise measurement of loss rates and is left for future work.
The experimental comparison is against direct teleoperation rather than against PID/MPC/NMPC, which are already done in Section~\ref{sec:simulation}.

\subsubsection{Test Vehicle}
A full-scale vehicle can capture real-world dynamics, but a 1/10th-scale platform is sufficient to demonstrate whether a controller is effective while offering a safer and more controllable environment. 
Consequently, we use an experimental vehicle on a 1/10-scale platform equipped with a Vedder Electronic Speed Controller (VESC) for direct motor torque control. 
This platform provides a safe, repeatable testbed for validating the proposed framework under realistic sensing/actuation delays and human-in-the-loop operation, without aiming to replicate full-scale vehicle dynamics.

\subsection{Experimental Results}
\subsubsection{Fundamental Longitudinal and Lateral Control Tests}
We first evaluate basic longitudinal and lateral responses to operator commands under both controlled (laboratory) and real-world conditions.

\textbf{Longitudinal tests:} Target speeds of 1~m/s and 2~m/s are tested (i) indoors without disturbances and (ii) outdoors under 5~m/s wind and uneven terrain. 
After the response stabilizes, we record the maximum speed deviation.

\textbf{Lateral tests:} Lane-change maneuvers are performed at 0.65~m/s and 1.2~m/s (i) indoors and (ii) outdoors on a 5$^\circ$ banked road. 
Performance is quantified by $W_l$ and $W_r$, defined as the distances from the wheels to the left/right lane borders after stopping, averaged over three trials (negative values indicate lane crossings).

\textbf{Baselines:} For longitudinal tests, we repeat the experiments without the proposed system. For lateral tests, we disable only the lateral module while keeping the RBFN-based speed controller active. 
The RV idle speed is adjusted to support autonomous regulation and reduce operator interference.

\subfile{tables/ad_ex_vel}

\subfile{pic/ad_vel_exe}
\subfile{tables/ad_ex_lateral_300.tex}
Table~\ref{Tab:ad_ex_vel_all} shows that the proposed system consistently reduces speed deviation in both laboratory and outdoor conditions. 
Deviations increase with higher target speed and latency, consistent with simulation in Section~\ref{sec:simulation}, but remain substantially smaller than without assistance. 
Figure~\ref{fig:ad_vel_exe} further shows that without the system the operator struggles to maintain the target speed—especially at 2~m/s—whereas with the system the RV tracks the target with minimal deviation.

Table~{\ref{Tab:ad_ex_lateral_all}} shows that without the proposed system, the operator loses lane containment as latency and speed increase. At 500\,ms latency and 1.20\,m/s on a banked road, the inner wheel crosses the lane border ($W = -0.15\,\mathrm{m}$ ): a safety failure. With the proposed system, the inner-wheel margin remains positive ($W_l \geq 0.2\,\mathrm{m}$ in all conditions tested), demonstrating that the framework is not merely improving accuracy but preventing a failure mode that direct teleoperation cannot avoid on its own.

\subsubsection{Track Driving with Desired Velocity}
We next evaluate closed-track driving when the operator specifies heading and desired speed.

We conduct two tests:
\begin{itemize}
  \item \textbf{Idle-speed test (0.65~m/s):} The RV maintains an idle speed, so the operator primarily controls heading. This tests heading controllability and idle-speed regulation.
  \item \textbf{Variable-speed test ([0.9, 1.2]~m/s):} The operator commands both heading and speed within a range, testing whether the system facilitates simultaneous speed and steering control.
\end{itemize}

For comparison, we repeat both tests without the proposed system. 
To keep the idle-speed condition comparable in the baseline, we use a PID controller to regulate the 0.65~m/s idle speed when the RBFN-based controller is removed. 
RV trajectories are recorded using a position sensor on the test platform.

\subfile{pic/vel_comp}
\subfile{pic/path_comp}

Figures~\ref{fig:experiment_result_vel} and~\ref{fig:experiment_result} summarize the results. 
In the idle-speed test, the proposed system maintains $v_f$ close to 0.65~m/s; the PID baseline can also regulate around this value, enabling a fair comparison of heading control. 
In the variable-speed test, the operator achieves smoother speed regulation with the proposed system: $v_f$ closely tracks the desired speed $v_d$ generated by the Trajectory Design Module, whereas the one without proposed system exhibits larger fluctuations.

Path results show that at low speed the operator can drive accurately even without assistance, but at higher speeds the baseline requires substantially more effort and exhibits degraded path tracking. 
With the proposed system, heading control remains smooth and the RV stays well within the track boundaries. 
These results also validate that the Trajectory Design Module generates feasible desired trajectories in real time, which is necessary for smooth teleoperation.

Finally, Figure~\ref{fig:experiment_result_delay} shows that the RV remains controllable under increasing latency. At lower speeds, trajectories remain smooth as delay increases; at higher speeds, control becomes more challenging with larger delay, but the RV still stays within the track, demonstrating the robustness of the proposed system.

\subfile{pic/delay_comp}

%% file: tables/ad_ex_vel.tex
\begin{table}[ht]
\centering
\scriptsize
\caption{Maximum Speed Deviation (m/s) Under Default and Elevated Time Latency Levels}
\begin{tabular}{m{1.2cm}<{\centering}|m{1.0cm}<{\centering}|m{1.0cm}<{\centering}m{1.0cm}<{\centering}|m{1.0cm}<{\centering}m{1.0cm}<{\centering}}
\toprule[1.3pt]
\midrule[0.3pt]
Latency & Speed & \multicolumn{2}{c|}{Lab Setting} & \multicolumn{2}{c}{Outside Road} \\
\cmidrule{3-6}
(ms) & (m/s) & with & without & with & without \\
\midrule
\multirow{2}{*}{Default} & 1 & 0.1 & 0.3 & 0.2 & 0.4 \\
\cmidrule{2-6}
& 2 & 0.2 & 0.5 & 0.4 & 0.6 \\
\midrule
\multirow{2}{*}{300} & 1 & 0.3 & 0.4 & 0.4 & 0.6 \\
\cmidrule{2-6}
& 2 & 0.4 & 0.7 & 0.6 & 1.1 \\
\midrule
\multirow{2}{*}{400} & 1 & 0.3 & 0.5 & 0.4 & 0.8 \\
\cmidrule{2-6}
& 2 & 0.4 & 0.7 & 0.6 & 1.2 \\
\midrule
\multirow{2}{*}{500} & 1 & 0.4 & 0.6 & 0.4 & 0.9 \\
\cmidrule{2-6}
& 2 & 0.5 & 0.8 & 0.6 & 1.3 \\
\midrule
\end{tabular}
\label{Tab:ad_ex_vel_all}
\end{table}

%% file: pic/ad_vel_exe.tex
\begin{figure}[!h]
    \centering
    \captionsetup{justification=justified}
    \includegraphics[width=8cm]{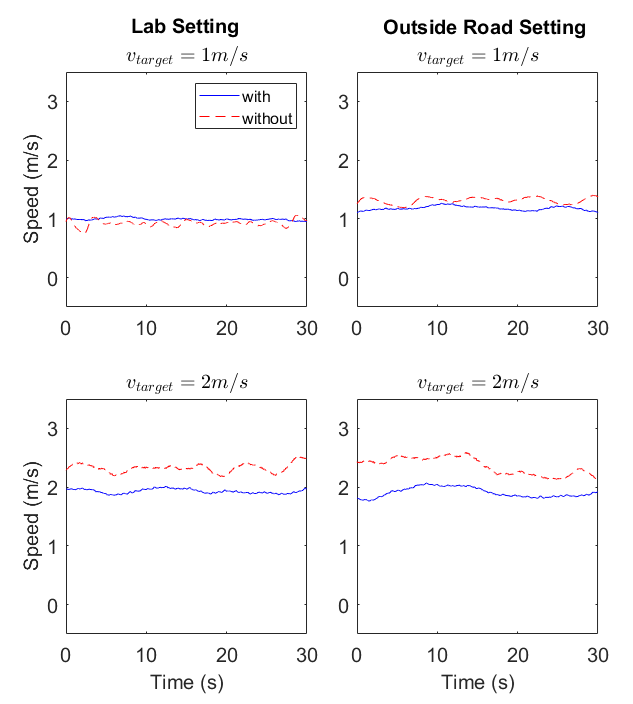}
    \caption{Longitudinal tracking performance under the default 4G latency setting.}
    \label{fig:ad_vel_exe}
\end{figure}

%% file: tables/ad_ex_lateral_300.tex
\begin{table*}[!h]
\centering
\scriptsize
\caption{Distance (m) between Wheel and Target Lane Border Line Under Default \\and Elevated Time Latency Levels}
\begin{tabular}{m{1.0cm}<{\centering}|m{1.0cm}<{\centering}|m{0.7cm}<{\centering}|m{0.7cm}<{\centering}|m{0.7cm}<{\centering}|m{0.7cm}<{\centering}|m{0.7cm}<{\centering}|m{0.7cm}<{\centering}|m{0.7cm}<{\centering}|m{0.7cm}<{\centering}|m{0.7cm}<{\centering}|m{0.7cm}<{\centering}|m{0.7cm}<{\centering}|m{0.7cm}<{\centering}}
\toprule[1.3pt]
\midrule[0.3pt]
Latency & Speed & \multicolumn{4}{c|}{Lab Setting} & \multicolumn{4}{c|}{Outside Road (Up)} & \multicolumn{4}{c}{Outside Road (Down)} \\ 
\cmidrule{3-14}
(ms) & (m/s) & \multicolumn{2}{c|}{with} & \multicolumn{2}{c|}{without} & \multicolumn{2}{c|}{with} & \multicolumn{2}{c|}{without} & \multicolumn{2}{c|}{with} & \multicolumn{2}{c}{without} \\                   
\cmidrule{3-14}
& & $W_l$ & $W_r$ & $W_l$ & $W_r$ & $W_l$ & $W_r$ & $W_l$ & $W_r$ & $W_l$ & $W_r$ & $W_l$ & $W_r$ \\                   
\midrule
\multirow{2}{*}{Default} & 0.65 & 0.31 & 0.34 & 0.33 & 0.32 & 0.30 & 0.35 & 0.25 & 0.40 & 0.32 & 0.33 & 0.40 & 0.25 \\ 

& 1.20 & 0.30 & 0.35 & 0.29 & 0.36 & 0.28 & 0.37 & 0.15 & 0.50 & 0.36 & 0.29 & 0.55 & 0.10 \\ 
\midrule
\multirow{2}{*}{300} & 0.65 & 0.30 & 0.35 & 0.30 & 0.35 & 0.28 & 0.37 & 0.20 & 0.45 & 0.32 & 0.33 & 0.40 & 0.25 \\ 

& 1.20 & 0.29 & 0.36 & 0.10 & 0.55 & 0.25 & 0.40 & 0.10 & 0.55 & 0.38 & 0.27 & 0.60 & 0.05 \\ 
\midrule
\multirow{2}{*}{400} & 0.65 & 0.29 & 0.36 & 0.25 & 0.40 & 0.28 & 0.37 & 0.00 & 0.65 & 0.33 & 0.32 & 0.41 & 0.24 \\ 

& 1.20 & 0.25 & 0.38 & -0.07 & 0.72 & 0.28 & 0.37 & -0.10 & 0.75 & 0.40 & 0.25 & 0.72 & -0.07 \\ 
\midrule
\multirow{2}{*}{500} & 0.65 & 0.27 & 0.38 & 0.25 & 0.40 & 0.27 & 0.38 & 0.20 & 0.45 & 0.37 & 0.28 & 0.42 & 0.23 \\ 

& 1.20 & 0.22 & 0.43 & -0.15 & 0.80 & 0.27 & 0.38 & -0.10 & 0.75 & 0.45 & 0.20 & 0.80 & -0.15 \\ 
\midrule
\end{tabular}
\label{Tab:ad_ex_lateral_all}
\end{table*}

%% file: pic/vel_comp.tex
\begin{figure}[h]
    \centering
    \captionsetup{justification=justified}
    \includegraphics[width=8.5cm]{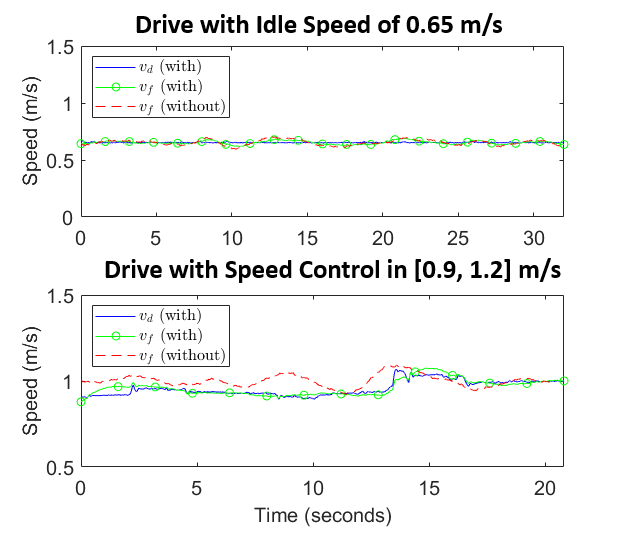}
    \caption{Vehicle's Velocity Performance with or without Designed System in the Experiment}
    \label{fig:experiment_result_vel}
\end{figure}

%% file: pic/path_comp.tex
\begin{figure}[h]
    \centering
    \captionsetup{justification=justified}
    \includegraphics[width=8.5cm]{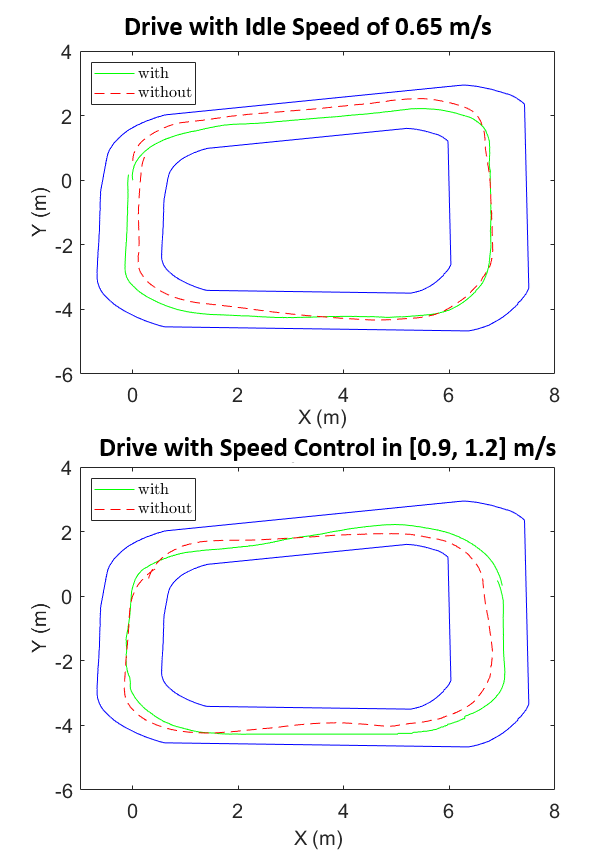}
    \caption{Recorded Vehicle Trajectory with or without Designed System in the Experiment}
    \label{fig:experiment_result}
\end{figure}

%% file: pic/delay_comp.tex
\begin{figure}[h]
    \centering
    \captionsetup{justification=justified}
    \includegraphics[width=8.5cm]{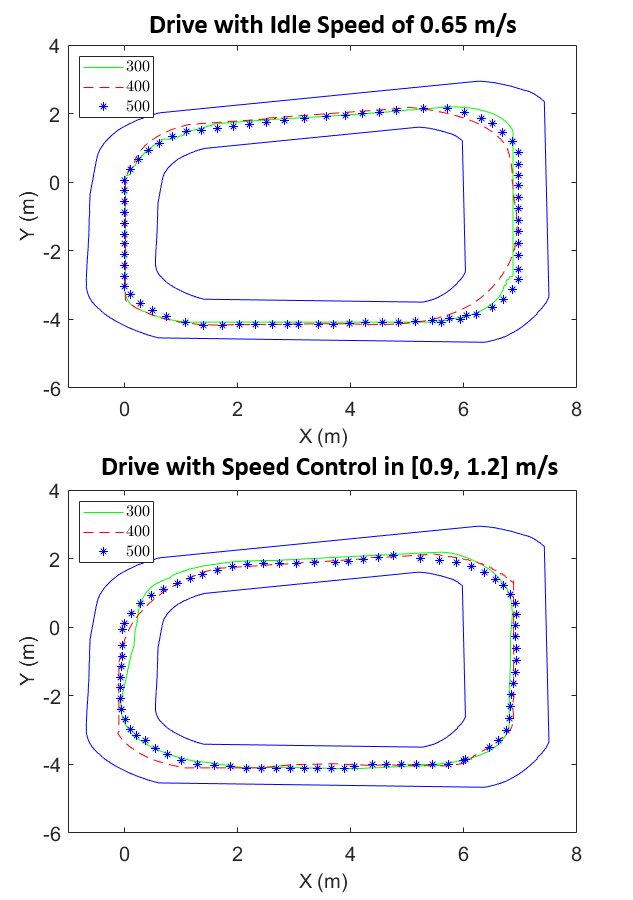}
    \caption{Recorded Vehicle Trajectory with Different Extended Time Latency }
    \label{fig:experiment_result_delay}
\end{figure}

%% file: CONCLUSIONS_and_FUTURE_WORK.tex
\section{CONCLUSIONS and FUTURE WORK}

This paper presents and validates a direct vehicle teleoperation system addressing communication latency and uncertainty. Under bounded signals, bounded RBFN basis functions, and a passive operator assumption, the WV channel preserves passivity and guarantees UUB tracking errors, while the RBFN compensates for uncertainties.

Simulation results show that the RBFN effectively handles abrupt environmental changes and runs orders of magnitude faster than MPC and NMPC, making it well suited to edge computing on vehicles. 
Hardware-in-the-loop experiments with a 1/10th scale vehicle over a 4G network further confirm safe and effective operation with a human operator. 

Across both studies, a single-layer RBFN provides sufficient approximation accuracy at ultra-low computational cost and supports a straightforward Lyapunov-based adaptive-law design.

Future work will extend the stability analysis to coupled MIMO vehicle dynamics, incorporate strategies for severe packet loss, and validate the approach on full-scale autonomous vehicle platforms.

%% file: Biography.tex
\subfile{bio/a}
\hfill

\subfile{bio/c}

%% file: bio/a.tex
\begin{wrapfigure}{l}{25mm} 

\includegraphics[width=1in,height=1.5in,clip,keepaspectratio]{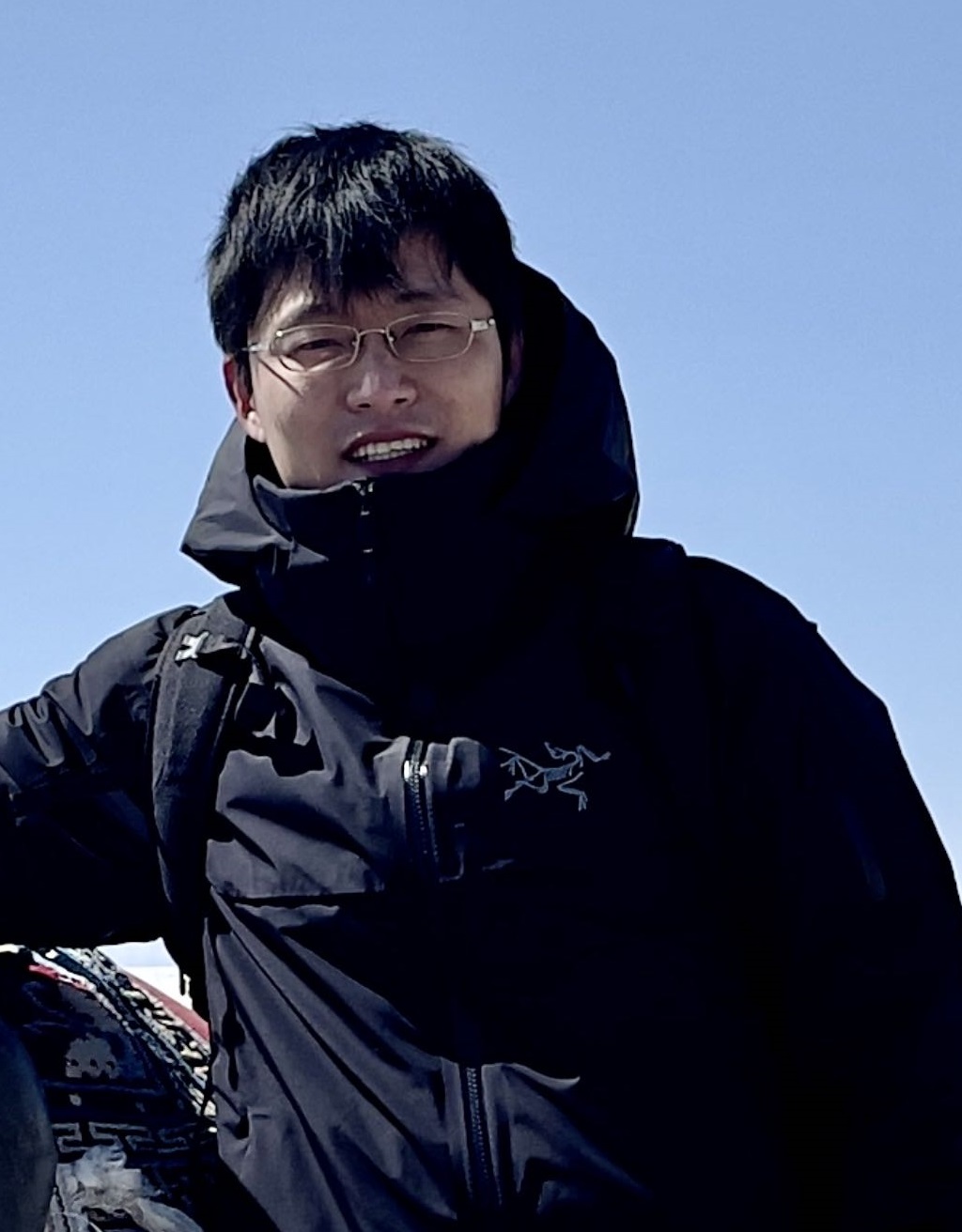}

\end{wrapfigure}\par

\textbf{Ning Ding} received his B.S. degree and M.S. degree, both in Vehicle Engineering (Automobile) from Tongji University, China. He received a Ph.D. degree in mechanical engineering at Virginia Tech Autonomous System and Intelligent Machines (ASIM) Lab. His research interests include autonomous vehicles, machine learning, human-machine interface.\par

%% file: bio/c.tex
\begin{wrapfigure}{l}{25mm}
\includegraphics[width=1in,height=1.5in,clip,keepaspectratio]{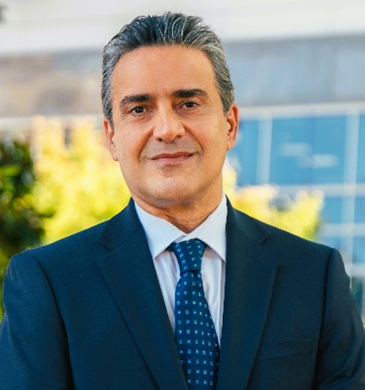}
\end{wrapfigure}\par
\textbf {Dr. Azim Eskandarian} has been Dean of the College of Engineering and Alice T. and William H. Goodwin Jr. Endowed Chair/Professor at Virginia Commonwealth University, Richmond, VA, since August 2023. Before that, he was a Professor and Head of the Mechanical Engineering Department at Virginia Tech since August 2015. He became the Nicholas and Rebecca Des Champs chair professor in April 2018 and a joint courtesy Electrical and Computer Engineering professor in 2021. He established the Autonomous Systems and Intelligent Machines laboratory at Virginia Tech, where he has conducted pioneering research in autonomous vehicles, human/driver cognition and vehicle interface, advanced driver assistance systems, and robotics. Before joining Virginia Tech, he was a Professor of Engineering and Applied Science at the George Washington University (GWU) and the Founding Director of the Center for Intelligent Systems Research from 1996 to 2015, the Director of the Transportation Safety and Security University Area of Excellence, from 2002 to 2015, and the Co-Founder of the National Crash Analysis Center in 1992 and its Director from 1998 to 2002 and 2013 to 2015. From 1989 to 1992, he was an Assistant Professor at Pennsylvania State University, York, PA, and an Engineer/Project Manager in the industry from 1983 to 1989. Dr. Eskandarian is a Fellow of ASME, a Fellow of IEEE (elevated in 2024), and a member of SAE professional societies. He received the SAE’s Vincent 2021 Bendix Automotive Electronics Engineering Award, the IEEE ITS Society’s Outstanding Researcher Award in 2017, and GWU’s School of Engineering Outstanding Researcher Award in 2013.
\par